\documentclass[lettersize,journal]{IEEEtran}
\usepackage{amsmath,amsfonts}
\usepackage{algorithmic}
\usepackage{algorithm}
\usepackage{array}
\usepackage[caption=false,font=normalsize,labelfont=sf,textfont=sf]{subfig}
\usepackage{textcomp}
\usepackage{stfloats}
\usepackage{url}
\usepackage{verbatim}
\usepackage{graphicx}
\usepackage{cite}
\usepackage{color}
\usepackage{booktabs}
\usepackage{amssymb}
\usepackage{mathtools}
\usepackage{amsthm}
\usepackage{wrapfig}
\usepackage{enumitem}
\usepackage{makecell}

\usepackage{listings}
\usepackage{xcolor}
\usepackage{adjustbox}

\newcommand{\review}[1]{\textcolor{black}{#1}}
\newcommand{\final}[1]{\textcolor{black}{#1}}

\newcommand{\onedot}{.}
\newcommand{\eg}{\emph{e.g}\onedot}

\newcommand{\ie}{\emph{i.e}\onedot}

\newcommand{\etal}{\emph{et al}\onedot}

\begin{document}
% \linenumbers
\title{Grammar-based Game Description Generation \\using Large Language Models}

\author{Tsunehiko Tanaka, Edgar Simo-Serra
        % <-this % stops a space
\thanks{The authors are with Waseda University, Tokyo, Japan. (Corresponding author: Tsunehiko Tanaka, email: tsunehiko@fuji.waseda.jp)}}

% The paper headers
% \markboth{Journal of \LaTeX\ Class Files,~Vol.~14, No.~8, August~2021}%
% {Shell \MakeLowercase{\textit{et al.}}: A Sample Article Using IEEEtran.cls for IEEE Journals}

% \IEEEpubid{0000--0000/00\$00.00~\copyright~2021 IEEE}
% % Remember, if you use this you must call \IEEEpubidadjcol in the second
% % column for its text to clear the IEEEpubid mark.

\maketitle

\begin{abstract}
\review{
Game Description Language (GDL) provides a standardized way to express diverse games in a machine-readable format, enabling automated game simulation, and evaluation.
While previous research has explored game description generation using search-based methods, generating GDL descriptions from natural language remains a challenging task.
This paper presents a novel framework that leverages Large Language Models (LLMs) to generate grammatically accurate game descriptions from natural language.
Our approach consists of two stages: first, we gradually generate a minimal grammar based on GDL specifications; second, we iteratively improve the game description through grammar-guided generation.
Our framework employs a specialized parser that identifies valid subsequences and candidate symbols from LLM responses, enabling gradual refinement of the output to ensure grammatical correctness.
Experimental results demonstrate that our iterative improvement approach significantly outperforms baseline methods that directly use LLM outputs. Our code is available at \url{https://github.com/tsunehiko/ggdg}}
\end{abstract}

\begin{IEEEkeywords}
Large Language Model, Ludii, Game Description Language, Grammar, Game Description Generation
\end{IEEEkeywords}

\section{Introduction}
\begin{figure}[t]
    \centering
    \includegraphics[width=0.9\linewidth]{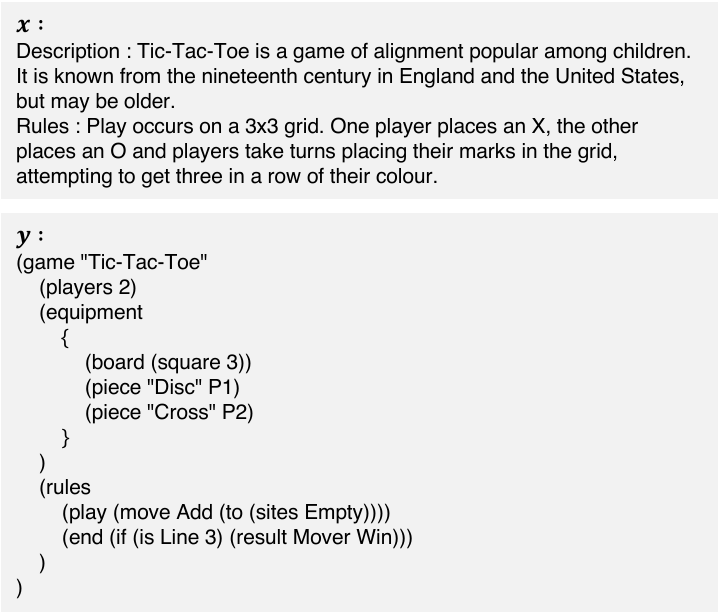}
    \caption{\textbf{An example of Ludii game description for the game ``Tic-Tac-Toe''.} $x$ is text that explains games in natural language. $y$ is a game description in Ludii GDL, a Game Description Language.}
    \label{fig:gdg_example}
\end{figure}

\review{
A Game Description Language (GDL)~\cite{gdl, vgdl, rbg, egd, ludii} is a domain-specific language that expresses a wide range of games in a unified notation.
For example, Ludii GDL~\cite{ludii} models over 1,000 games, primarily board games, as shown in Fig.~\ref{fig:gdg_example}.
Game descriptions represented in GDLs are highly machine-readable, making it easy to simulate gameplay using dedicated game engines.
Given the amenability of GDLs for automatic game evaluation, they have been extensively used in research on automated game design.
In particular, search-based methods such as evolutionary algorithms~\cite{egd}, MCTS~\cite{arcade, rule_generation}, and random forests~\cite{random_forest} have proven successful in generating game descriptions.
Most research primarily focused on mutating existing games based on fitness functions to generate novel games.
However, the task of generating game descriptions from natural language texts has not yet been sufficiently explored, and has the potential to lower the bar of entry to game design to non-specialists.
In this research, we use Large Language Models (LLMs)~\cite{llama3, gpt4}, which excel at understanding textual context, to generate game descriptions from natural language text in a two-stage process to enforce grammatical correctness.}

\review{
LLMs are language models with an enormous number of parameters, pre-trained on vast amounts of text data.
These models possess the ability to solve various tasks without additional training~\cite{program_synthesis_0, program_synthesis_1}, and this ability can be elicited by including the task context in the prompt, a technique known as In-Context Learning (ICL)~\cite{icl}.
Hu \etal~\cite{llmgg} applied this capability to game description generation by incorporating explanations for GDL notations and examples of game descriptions in the prompt context.
Their results have shown that more accurate game descriptions can be generated by appropriately refining the prompt context.
However, LLMs may still generate grammatically incorrect game descriptions.
Such grammatically inaccurate game descriptions cannot be correctly parsed and loaded by game engines, making it difficult to evaluate them through gameplay simulation.
To generate higher-quality games, it is important first to ensure that LLMs can produce grammatically correct GDL game descriptions.}

\review{
This paper presents a method for LLMs to generate more grammatically accurate game descriptions.
We propose an approach to iteratively improve LLMs' initial responses using the GDL grammar.
Our generation framework consists of two stages: (i) generating the minimal grammar required for game descriptions, and (ii) iteratively improving the game description based on this minimal grammar.
First, we use LLMs to generate the minimal grammar required to produce the game description, making use of the GDL grammar.
Next, a parser based on the minimal grammar determines grammatically valid subsequences and a set of candidate symbols that could follow these subsequences from the LLMs' responses.
LLMs then re-infer the missing parts based on these subsequences and candidate symbols, gradually generating grammatically accurate game descriptions.
Experimental results demonstrate that our framework is more effective in generating game descriptions compared to a baseline that directly uses the LLMs' initial responses as an output.}

Our contributions can be summarized as follows:
\begin{itemize}
\item We propose a framework for \review{generating game descriptions from natural language text} by using LLMs and GDLs.
\item \review{Our framework incorporates GDL grammar into the generation process and iteratively improves the grammatical correctness of the LLM's output.}
\item We propose iterative improvement decoding methods specialized for grammar generation and game description generation, respectively.
\item We demonstrate the effectiveness of our framework through extensive experiments on game description generation.
\end{itemize}

\begin{figure*}[t]
    \centering
    \includegraphics[width=\linewidth]{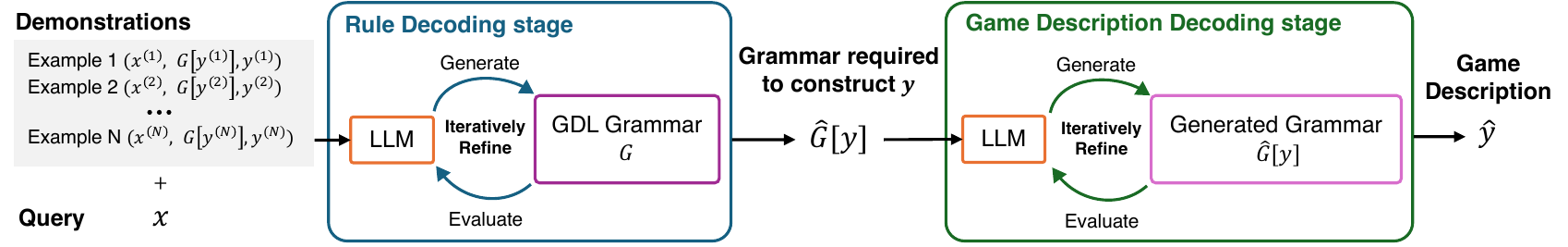}
    \vspace{1pt}
    \caption{\textbf{An overview of our framework for grammar-based game description generation.}
    \review{We generate the game description $y$ from a natural language query $x$ using Large Language Models (LLMs).}
    First, we generate the grammar $\hat{G}[y]$ required to construct $y$ in the Rule Decoding stage, and then generate $\hat{y}$ based on $\hat{G}[y]$ in the Game Description Decoding stage.
    \review{The core of our framework is iteratively decoding by leveraging the grammar of game description languages (GDLs) to improve the initial response from LLMs.
    We make use of Ludii GDL as our GDL, which can model a larger variety of games and is a context-free grammar.}}
    \label{fig:teaser}
\end{figure*}

\section{Related Work}
\subsection{Game Description Language}
A Game Description Language (GDL) is a domain-specific language for games.
\review{GGP-GDL}~\cite{gdl} was introduced in 2005 and has become the standard in General Game Playing, where artificial agents are developed to play a wide variety of games.
After \review{GGP-GDL}, various types of GDLs have been developed.
VGDL~\cite{vgdl} is a language that represents both the levels and rules of 2D sprite-based games, \review{and it models 195 games.}
RBG~\cite{rbg} models complex board games by combining low-level and high-level languages.
Ludii~\cite{ludii} is a game system developed based on the ludemic approach, which decomposes games into conceptual units of game-related information.
Ludii models more than 1,000 traditional games, including board games, card games, dice games, and tile games.
\review{Ludii's grammar is a Context-Free Grammar (CFG) in Extended Backus-Naur Form (EBNF) style~\cite{ebnf}.
Due to its ability to generalize and model more games, as well as its capacity to express complete game descriptions in CFG, we use Ludii GDL as our GDL.
Note that our approach can be applied to other GDLs that follow CFG, \eg, the rule section of VGDL.}

Diverse analyses of games using Ludii, especially focusing on board games, have been conducted~\cite{measuring, manual, board_concept, heuristic, universal}.
\review{For example, the distance between board games using concept values extracted from Ludii has been proposed in \cite{measuring}.}
\review{Stephenson} \emph{et al.}~\cite{manual} have presented a framework for automatically generating board game manuals using Ludii.
Unlike these works, we add a new perspective to the analysis by focusing on game description generation using grammar.

\subsection{Automated Game Design}
Automated game design~\cite{automatic, evolutionary} is one of the core themes in the field of game AI.
As AI technology rapidly advances, various aspects of how AI can be utilized in automated game design have been discussed, including the design process~\cite{aibased}, design patterns~\cite{pattern}, and creative machine learning~\cite{gdcml}.
\review{Deep learning-based ML is often used in the domain of procedural content generation for level design~\cite{levelllm, shyam2023mariogpt, chatgpt4pcg, pcgml, dlpcg, pcgrl}.}
Several studies~\cite{angelina, orchestrating} propose methods to generate a game by integrating multiple elements such as visuals, audio, narrative, levels, rules, and gameplay.
Generating games using new representations such as answer set programming~\cite{asp} and game graphs~\cite{expansion} \review{has also been explored}.
\review{Word2World~\cite{word2world} uses LLMs to design from stories to playable games procedurally.
These approaches design games without including GDLs.}

Automated game design for generating game designs in the \review{GDL} format has been explored~\cite{egd, arcade, cicero, rule_generation, random_forest}.
\review{The Ludi system~\cite{egd}, which was the precursor to the ludemic approach used in later Ludii~\cite{ludii}, employed evolutionary game design and generated the commercially viable game Yavalath.}
Thorbjørn \emph{et al.}~\cite{arcade} explored the approach to generate VGDL~\cite{vgdl} for arcade games using evolutionary algorithms.
GVG-RG~\cite{rule_generation} proposes a framework for generating appropriate game rules for given game levels.
Cicero~\cite{cicero} is a mixed-initiative tool that assists in prototyping 2D sprite-based games using VGDL.
Thomas \emph{et al.}~\cite{random_forest} aimed to acquire a fitness function that guides game design generation using adversarial random forest classifiers.
\review{These approaches generate novel games based on existing games, but our approach differs in that it generates game descriptions from natural language text.}

\subsection{Large Language Models in Games}
Since the advent of ChatGPT~\cite{openai_2022} in late 2022, Large Language Models (LLMs) have attracted significant attention, and various ways of using LLMs in \review{automated game design} \review{have been explored~\cite{gallotta2024large, practicalllm}}.
\review{
Several studies~\cite{levelllm, shyam2023mariogpt} have fine-tuned GPT-2~\cite{gpt2} models to generate 2D tile-based game levels.
Prompt-based approaches for LLMs have also been proposed for level generation~\cite{chatgpt4pcg}.
Other researchers have trained GPT models to generate quests for role-playing games~\cite{rpg}.
Dreamcraft~\cite{dreamcraft} is a method that uses LLMs to generate 3D game objects for Minecraft from text prompts.
In addition to generation, researchers have focused on evaluating LLMs.
Studies have assessed how well prompt-based level generation methods replicate and generalize~\cite{prompt-wrangling}, and have evaluated the capabilities of LLMs in mixed-initiative game design~\cite{co-creative}.
}

\final{GAVEL~\cite{gavel} uses LLMs fine-tuned on Ludii’s game descriptions as mutation operators in evolutionary search. While GAVEL aims to generate games with high novelty, our goal is to generate game descriptions that align with natural language text.}
LLMGG~\cite{llmgg} generates both the rules and levels of games represented in VGDL using LLMs.
In LLMGG, LLMs take a part of the VGDL~\cite{vgdl} representation or examples of other games as prompts, which generate a complete VGDL-based game in one step.
The authors discuss that incorrect game levels and rules not included in the VGDL grammar are generated.
In contrast, our approach involves multi-step generation, and the necessary grammar rules to build a game \review{description} are generated as intermediate representations in the middle steps.
\review{Moreover, since the game description is iteratively generated based on these grammatical rules, this approach prevents inaccurate syntax and improves consistency.}

\subsection{Program Synthesis}
In program synthesis, the task of generating programs from natural language is called semantic parsing.
Semantic parsing has benefited from advancements in LLMs.
Several efforts~\cite{program_synthesis_0, program_synthesis_1} have already explored generating code in general-purpose programming languages such as Python using LLMs.
To improve the accuracy of the generated programs, constrained decoding~\cite{shin-etal-2021-constrained, Scholak2021:PICARD, poesia2022synchromesh} has also been studied.
\review{
Grammar-constrained decoding~\cite{gcd, gad, syncode} restricts the output space of LLMs to a space described by a grammar, enabling the generation of structured outputs like programs.
Our framework includes decoding for grammar generation, in addition to grammar-constrained decoding for game descriptions.}

\review{Evolutionary algorithms that use LLMs as evolutionary operators~\cite{evolutionLM, eureka, sga, prompt_evolution} are also gaining attention in program synthesis.
Quality-diversity algorithms utilizing LLMs have been proposed for controllable program synthesis, such as neural architecture search~\cite{llmatic}.
These studies open up the possibility that using LLMs can also improve GDL generation.}

In domain-specific language (DSL) generation, a prompting method~\cite{grammar_prompting} has been proposed that introduces grammar as an intermediate product while LLMs iteratively reason to solve tasks, also known as a chain-of-thought reasoning~\cite{chain_of_thoughts}.
However, in \cite{grammar_prompting}, the evaluation targets simple and short DSLs such as SMCalFlow~\cite{smcalflow} and GeoQuery~\cite{geoquery}, and the use of LLMs in the approach is limited.
We aim to make more effective use of LLMs in our approach and evaluate it with complex and lengthy DSLs that represent game design, such as Ludii~\cite{ludii}.

\section{\review{Problem Setting}}

\begin{figure}[t]
    \centering
    \includegraphics[width=0.9\linewidth]{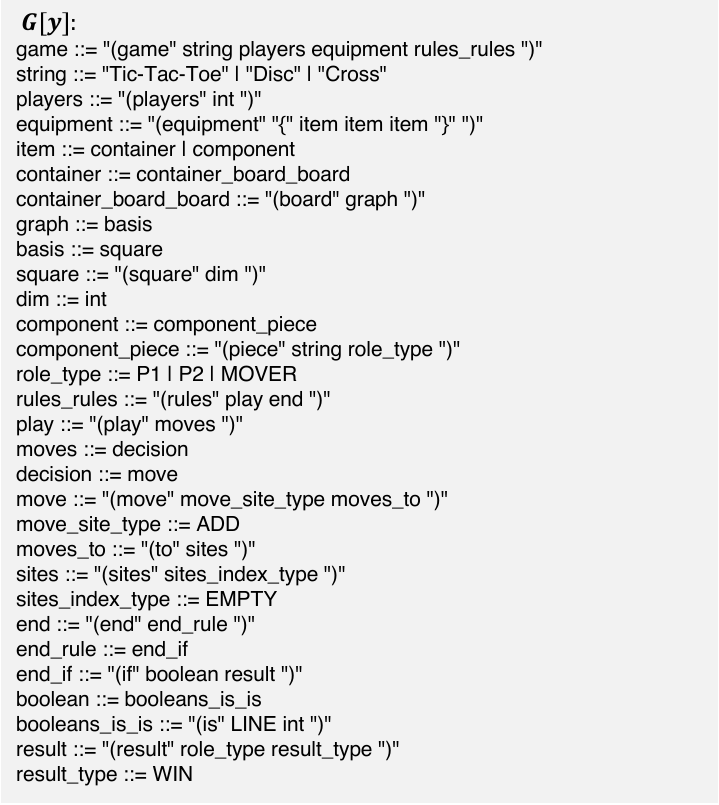}
    \caption{\textbf{Minimal Backus-Naur Form (BNF) grammar $G[y]$ for Tic-Tac-Toe.} Redundant rules are omitted for simplicity.}
    \label{fig:tic-tac-toe_grammar}
\end{figure}

\begin{table*}[t]
    \centering
    \caption{\review{\textbf{Glossary of notations used in this paper}.}}
    \begin{tabular}{ll}
    \toprule
    Notation & Description \\
    \midrule
    $G$ & GDL grammar (Ludii grammar in this paper) \\
    $L(G)$ & Set of game descriptions generated by $G$ \\
    $x$ & Texts explaining the rules of a game \\
    $y$ & A game description (ground truth) \\
    $G[y]$ & Minimal grammar extracted from $G$ containing only the grammar rules necessary to generate game description $y$ \\
    $(x^{(i)}, y^{(i)})_{I=1}^N$ & Demonstrations of game description generation to be included in LLMs' prompts \\
    $\hat{G}[y]_\mathrm{valid}$ & Set of grammar rules in the predicted $\hat{G}[y]$ that are included in $G$ \\
    $N_U$ & Set of non-terminal symbols defined in $G[y]$ but not defined in $\hat{G}[y]_\mathrm{valid}$ \\
    $G_{N_U}$ & Set of grammar rules in $G$ defining $N_U$ \\
    $G[y]_{N_U}$ & Minimal set of rules extracted from $G_{N_U}$ necessary to generate $y$ \\
    $\hat{y}_\mathrm{valid}$ & Part of $\hat{y}$ that conforms to $\hat{G}[y]$ \\
    $\Omega_\mathrm{candidate}$ & Set of candidate terminal symbols to follow $\hat{y}_\mathrm{valid}$ \\
    $\omega$ & Terminal symbol selected from $\Omega_\mathrm{candidate}$ by generators such as LLMs \\
    $\hat{\cdot}$ & All hat notations indicate symbols are predictions \\
    $\cdot^{\prime}$ & All prime notations indicate that the symbols are in the LLM's output space. \\
    & Among these, the one that maximizes the generation probability receives the hat notation. \\
    \bottomrule
  \end{tabular}
  \label{tab:glossary}
\end{table*}

In this section, we define our problem setting.
We first describe the generation of games written in GDLs.
We next review in-context learning in the game description generation.
\review{Notations used in this paper are summarized in Tab.~\ref{tab:glossary}.}

\subsection{Game Description Generation}
\review{Let $G$ be the GDL grammar, and let $L(G)$ represent the set of game descriptions generated by $G$.
We call the task of inputting a natural language query $x$ that describes the content and rules of a game, and generating a corresponding game description $y \in L(G)$, \textit{game description generation}.
Our goal is to make the generated game description $\hat{y}$ as close as possible to the ground truth $y$.}

\review{
In this paper, we use Ludii GDL as our GDL as it is one of the main references in GDL research.
Additionally, Ludii's grammar is a CFG in EBNF style, which allows us to generate complete game descriptions based on a CFG.
%CFG has a wide range of applications, and since our approach, which we will explain later, is based on CFG, this property of Ludii is beneficial to us.
An example of Ludii game description generation is shown in Fig.~ \ref{fig:gdg_example}.
From this point forward, unless otherwise specified, $G$ represents Ludii’s grammar.}

\subsection{In-Context Learning}
In-Context Learning (ICL)~\cite{icl} is an efficient method that provides pretrained LLMs with a few task-specific examples to obtain more precise and accurate results.
This approach does not require additional training or fine-tuning; instead, it relies on the LLM's ability to identify and apply patterns from the provided examples.
In ICL, LLM is conditioned on $N$ demonstrations $(x^{(i)}, y^{(i)})_{i=1}^{N}$ followed by a test example query $x$, and generates $y$ as $P_\mathrm{LLM}(y|(x^{(i)}, y^{(i)})_{i=1}^N, x)$.
Recent studies~\cite{pal,chain} have reported that the few-shot performance on complex reasoning tasks can be improved by inserting intermediate reasoning steps between $x^{(i)}$ and $y^{(i)}$ in the demonstrations.

The effectiveness of ICL depends on how effectively the solution to a task can be conveyed through demonstrations.
Intuitively, providing more demonstrations to the LLM seems to be beneficial.
However, the context length, which is the maximum length that the LLM can capture, is determined during pre-training.
Therefore, it is not possible to input a number of demonstrations that exceed this context length.
For example, when applying Llama3~\cite{llama3}, one of the leading open-source LLMs, to Ludii's game descriptions, the context length can easily be exceeded.
In particular, the average token length of Ludii's game descriptions using the Llama3-8B-Instruct tokenizer is 2,458.
The context length of Llama3-8B-Instruct is 8,192, which limits the number of Ludii demonstrations that can be input to a few at most.
This limitation when using complex examples such as those described in Ludii is common to many LLMs.
\review{Game descriptions generated based on such limited context lack grammatical accuracy and cannot make the games functional and will be discussed in Sec.~\ref{sec:experimental_results}.}

\section{Methodology}
\begin{figure*}[t]
    \centering
    \includegraphics[width=0.96\linewidth]{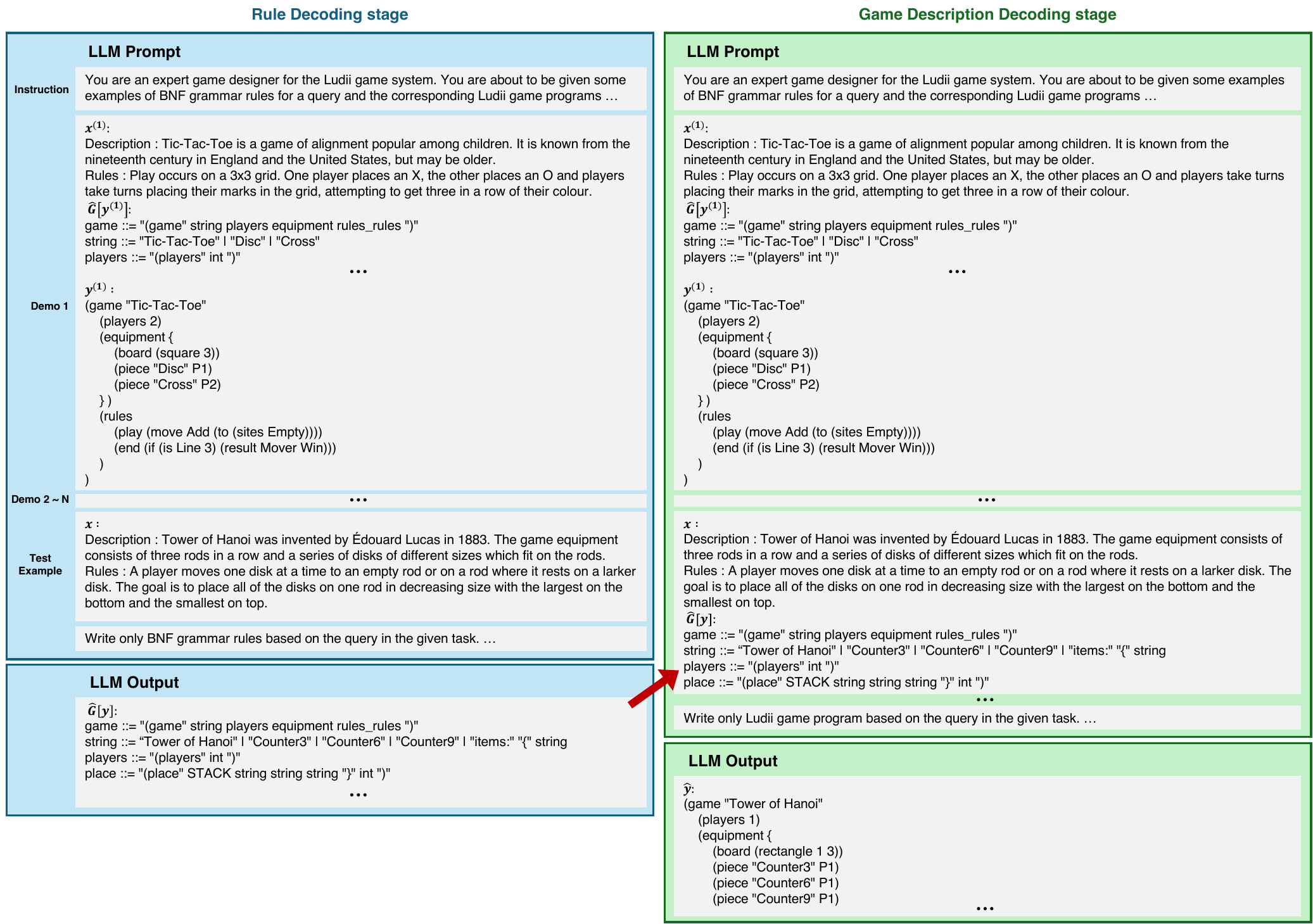}
    \caption{\textbf{An example of grammar-based game description generation result.}
    We generate game descriptions in two stages: first we generate the required grammar, and then generate the game description based on the grammar.
    The prompt includes demonstrations for in-context learning and the test example query $x$, and the demonstration contains the grammar $G[y^{(i)}]$.
    In the first stage, the minimal grammar $\hat{G}[y]$ that composes $y$ is generated.
    In the second stage, game description $\hat{y}$ is generated based on the generated $\hat{G}[y]$. The red arrow indicates that $\hat{G}[y]$ generated in the first stage is included in the prompt for the second stage.}
    \label{fig:twostage_generation}
\end{figure*}
\review{
In this section, we explain our approach to iteratively improve the initial responses of the LLM.
To evaluate the game description and improve its grammatical accuracy, we introduce the minimum grammar \( G[y] \subseteq G \) required to construct \( y \).
This minimum grammar \( G[y] \) is expected to provide more context to LLMs when included in the prompt, thereby enhancing In-Context Learning (ICL).
As shown in Fig.~\ref{fig:teaser}, our generation process consists of two stages:
First, we input the query \( x \) of the test example into an LLM to generate the minimum grammar \( \hat{G}[y] \) required to construct \( y \).
\( \hat{G}[y] \) is then evaluated, and the parts that conform to the Ludii grammar \( G \) are extracted.
The LLM then generates the missing rules.
Next, based on \( \hat{G}[y] \), the LLM generates the game description \( \hat{y} \).
\( \hat{y} \) is evaluated, and the parts that conform to \( \hat{G}[y] \) are extracted.
The LLM then infers the rest. Evaluation and LLM generation are repeated at each stage.
In this section, we first introduce the minimum grammar \( G[y] \), then explain the two-stage generation, and finally describe the decoding process for each stage in detail.
}

\subsection{Grammar-based Game Description Generation}
\label{subsec:generation}
\review{$G[y]$ is the minimal grammar that extracts only the rules necessary to generate $y$ from all the rules of $G$.
An example of $G[y]$ for tic-tac-toe is shown in Figure \ref{fig:tic-tac-toe_grammar}. }
$G[y]$ is a subset of the full grammar $G$, where $y \in L(G[y])$ and $\forall{r} \in G[y], y \notin L(G[y] \setminus \{r\})$.
\review{For any rule $r$ in $G[y]$, removing $r$ makes it impossible to generate $y$ ($y \notin L(G[y] \setminus {r})$). 
$G[y]$ is minimal in the sense that it contains exactly the rules required to generate $y$, with no superfluous rules.
A parser based on this minimal grammar can determine grammatically valid subsequences and the set of candidate symbols that follow them from the LLM’s responses.
The LLM then generates the rest of the game description based on the subsequences and candidate symbol groups.
Intuitively, it is expected that repeating this parsing and generation process will result in a more grammatically accurate game description than the initial response from the LLM.
This iterative improvement method is explained in Sec.~\ref{subsec:decoding}.}

\review{From the perspective of ICL, providing more context to LLMs is expected to have a positive impact.
It is known that providing GDL grammar to LLMs can improve the quality of generated game descriptions, especially in terms of grammatical accuracy~\cite{llmgg}.
However, the token length of Ludii's grammar using the Llama3-8B-Instruct tokenizer is 15,442, which is longer than the context length of most LLMs.
The straight-forward approach of including all the GDL grammar in the prompt as proposed in LLMGG is not feasible without significant more computation power.
The average token length of $G[y]$ using the Llama3-8B-Instruct tokenizer is 1,031, which can be added to the prompt within the context length limit for several games.}
In this case, each demonstration consists of $(x^{(i)}, G[y^{(i)}], y^{(i)})$.
$G[y^{(i)}]$ is obtained by parsing $y^{(i)}$ with $G$ and collecting the rules necessary to derive $y^{(i)}$.

The generation process consists of two stages.
In the first stage, a few demonstration examples and a test example query $x$ are input to the LLM as a prompt to generate the minimal grammar $\hat{G}[y]$ necessary to compose $y$.
The grammar $G'$ that maximizes the following probability is selected as $\hat{G}[y]$,
\begin{align}
    \label{eq:rule_generation}
    P_\mathrm{LLM}(G'|(x^{(i)}, G[y^{(i)}], y^{(i)})_{i=1}^N, x).
\end{align}
In the second stage, the generated $\hat{G}[y]$ is added to the prompt to generate the game description $\hat{y}$, and the $y'$ that maximizes the following probability is selected as $\hat{y}$:
\begin{align}
    \label{eq:description_generation}
    P_\mathrm{LLM}(y'|(x^{(i)}, G[y^{(i)}], y^{(i)})_{i=1}^N, x, \hat{G}[y]).
\end{align}
We show an overview of the two-stage generation process in Fig.~\ref{fig:twostage_generation}.

Additionally, in the second stage, although $y$ is conditioned on the grammar $\hat{G}[y]$, the generated game description may not adhere to $\hat{G}[y]$.
This is because the LLM just selects the words with the highest likelihood, and does not necessarily comply with the conditions set by the prompt.
Similarly, $\hat{G}[y]$ may not be a subset of the original grammar $G$.
Therefore, in the next subsection, we propose decoding methods to improve the consistency of the minimal grammar $\hat{G}[y]$ with the original grammar $G$, and the consistency of the game description $\hat{y}$ with the grammar $\hat{G}[y]$.

\begin{figure*}[t]
    \begin{minipage}[b]{0.475\textwidth}
        \centering
        \includegraphics[width=\linewidth]{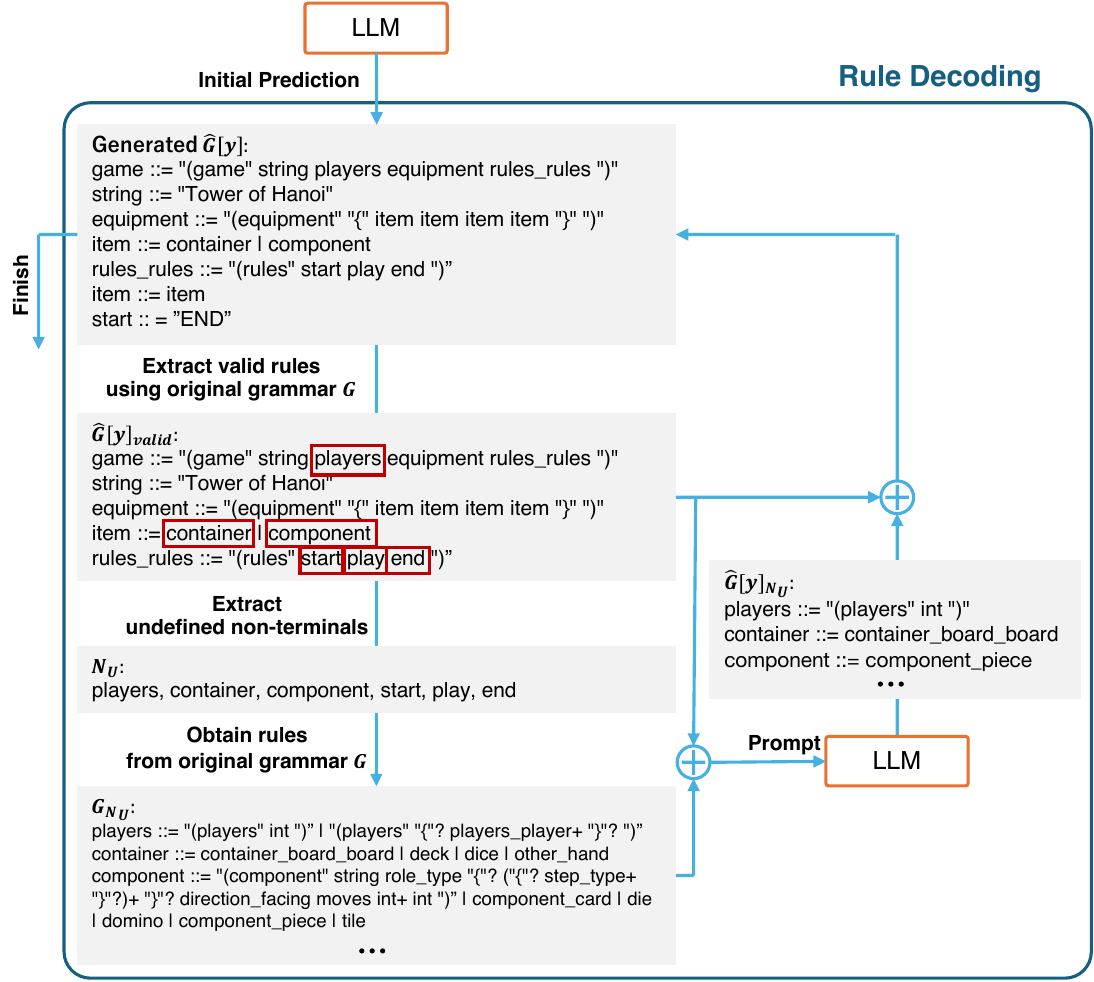}
        \caption{\textbf{Processing flow of Rule Decoding stage.}
        The Rule Decoding stage starts from the minimal grammar $\hat{G}[y]$ necessary to compose $y$ generated by the LLM, and improves it iteratively.
        From the grammar $\hat{G}[y]$, the set of rules included in the original grammar $G$ is extracted as $\hat{G}[y]_\mathrm{valid}$.
        Next, undefined non-terminal symbols $N_U$ are extracted from $\hat{G}[y]_\mathrm{valid}$.
        The rules $G_{N_U}$ concerning $N_U$ are obtained from the original grammar $G$ and input to the LLM along with $\hat{G}[y]_\mathrm{valid}$.
        The LLM then generates rules $\hat{G}{N_U}$ for the undefined non-terminal symbols.
        Finally, $\hat{G}[y]$ is updated by combining $G_{N_U}$ with $\hat{G}[y]_\mathrm{valid}$.}
        \label{fig:rule_decoding}
    \end{minipage}
    \hspace{0.025\textwidth}
    \begin{minipage}[b]{0.475\textwidth}
        \centering
        \includegraphics[width=0.9\linewidth]{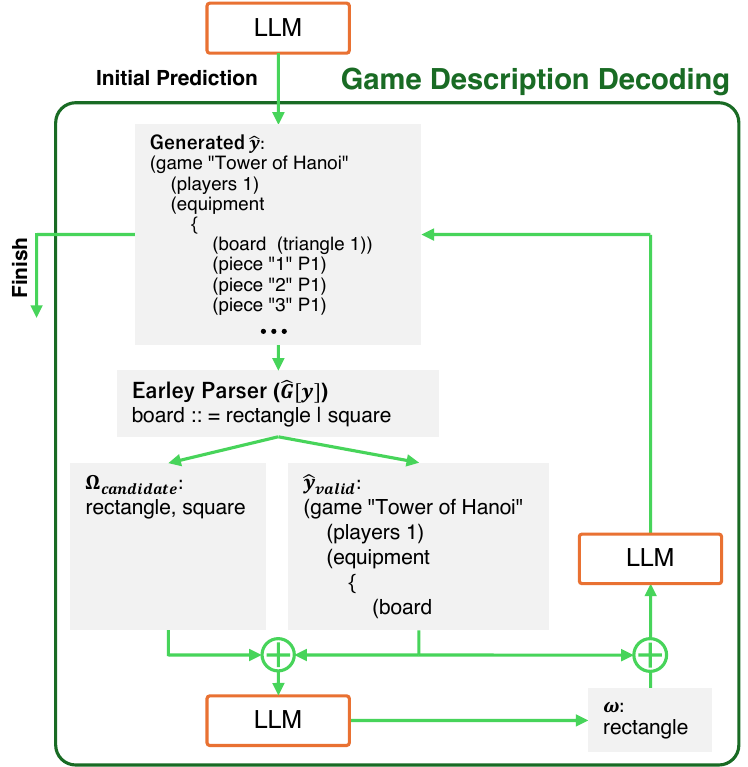}
        \caption{\textbf{Processing flow of Game Description Decoding stage.}
        The Game Description Decoding stage starts from the initial game description $\hat{y}$ generated by the LLM, and then enhances the results iteratively.
        Using an Earley parser~\cite{earley} based on $\hat{G}[y]$, the longest valid subsequence $\hat{y}_\mathrm{valid}$ and the subsequent terminal symbol candidates $\Omega_\mathrm{candidate}$ are obtained.
        The LLM then selects the candidate $\omega$ from $\Omega_\mathrm{candidate}$ that is most suitable as the successor to $\hat{y}_\mathrm{valid}$.
        $\omega$ is appended to the end of $\hat{y}_\mathrm{valid}$, and the LLM generates the remaining part of the game description $\hat{y}$ that follows.
        The generated $\hat{y}$ replaces the previous $\hat{y}$ from the earlier step.}
        \label{fig:game_description_decoding}
    \end{minipage}
\end{figure*}

\subsection{Grammar-based Iterative Decoding}
\label{subsec:decoding}
In order to overcome issues of inconsistent grammars, we propose decoding methods that iteratively improve the generated grammars.
In particular, our decoding methods are divided into two types, one specialized for the grammar $\hat{G}[y]$, and one for the game description $\hat{y}$.
When decoding the grammar $\hat{G}[y]$, undefined non-terminal symbols are extracted in one step, and rules to define them are then generated in the next step, in what we call the \textit{Rule Decoding} stage.
Similarly, when generating the game description $\hat{y}$, we use the Earley parser~\cite{earley} to obtain the longest valid continuation that adheres to the grammar $\hat{G}[y]$ in one step, and then complete the remaining parts following the valid continuation in the next step.
We refer to this as \textit{Game Description Decoding} stage.

\paragraph{Rule Decoding Stage}
Our rule decoding stage iteratively applies rule decoding while aiming to ensure that the generated $\hat{G}[y]$ is a subset of the original grammar $G$.
In our rule decoding stage, the generation process is multi-step, where each step iteratively improves the grammar $\hat{G}[y]$.
The goal of each step is to define the non-terminal symbols that were not defined in the grammar $\hat{G}[y]$ generated in the previous step.
We illustrate our rule decoding process in Fig.~\ref{fig:rule_decoding}.
Initially, $\hat{G}[y]$ is generated using the LLM in the same way as in Eq.~\eqref{eq:rule_generation}.
From $\hat{G}[y]$, only the valid grammar rules $\hat{G}[y]_\mathrm{valid}$ are retained, which are included in the original grammar $G$.
Among $\hat{G}[y]_\mathrm{valid}$, the set of undefined non-terminal symbols $N_U$ is extracted.
Undefined non-terminal symbols are rules that are used on the right-hand side of rules in $\hat{G}[y]_\mathrm{valid}$ but are not defined. 
They can be automatically and easily extracted by identifying symbols not appearing on the left-hand side of any rules.
The set of grammar rules for the undefined non-terminal symbols $G_{N_U}$ is extracted from the original grammar $G$.
$G_{N_U}$ is added to the prompt, and the LLM selects only the necessary rules from $G_{N_U}$ based on the query $x$.
\review{The right-hand side of the rules in $G_{N_U}$ includes options that are unnecessary for constructing $y$.
The role of the LLM is to select the minimum necessary choices to construct $y$ from these options, and to predict the minimal grammar $G[y]_{N_U} = G[y] \setminus \hat{G}[y]_\mathrm{valid}$.}
% Since $G_{N_U}$ is extracted from the original grammar $G$, it contains more rules than the minimum required for $y$.
% The role of the LLM is to extract only the minimal rules necessary to compose $y$ from $G_{N_U}$.
The rules that maximize the following probability are selected:
\begin{align}
    P_\mathrm{LLM}(G'[y]_{N_U} |(x^{(i)}, G[y^{(i)}], y^{(i)})_{i=1}^N, x, \hat{G}[y]_\mathrm{valid}, G_{N_U}).
\end{align}
The generated $\hat{G}[y]_{N_U}$ is combined with $\hat{G}[y]_\mathrm{valid}$ to update $\hat{G}[y]$.
This process is repeated until there are no more non-terminal symbols or a predetermined limit of updates is reached.

\begin{figure*}[t]
    \centering
    \includegraphics[width=\linewidth]{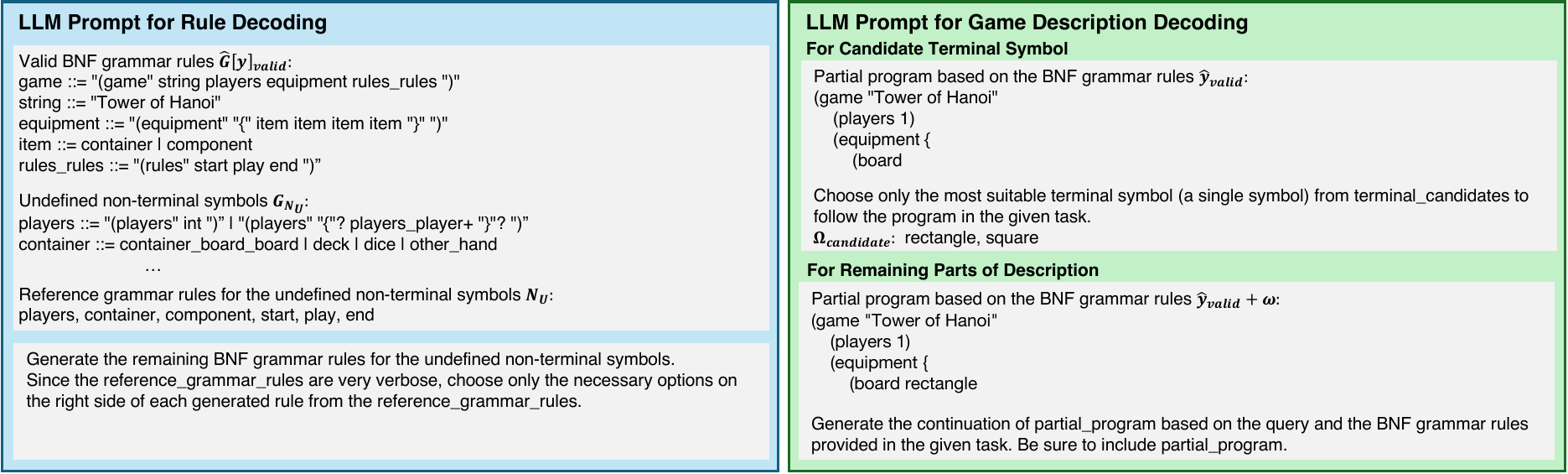}
    \caption{\textbf{Our prompts for grammar-based iterative decoding.}}
    \label{fig:prompts}
\end{figure*}

\paragraph{Game Description Decoding stage}
The game description decoding stage consists of iterative application of game description decoding to generate game descriptions that more accurately adhere to the grammar $\hat{G}[y]$.
In each step, it aims to complete the parts of $\hat{y}$ generated in the previous step that cannot be parsed.
Initially, $\hat{y}$ is generated using the LLM in the same way as in Eq.~\eqref{eq:description_generation}.
An Earley parser~\cite{earley} is then employed using $\hat{G}[y]$.
The Earley parser explores the program from left to right, extracting the longest valid subsequence and the subsequent candidate terminal symbols.
Using the generated $\hat{y}$, the Earley parser obtains the valid subsequence $\hat{y}_\mathrm{valid}$ and the set of candidate terminal symbols $\Omega_\mathrm{candidate}$.
\final{Because multiple options for rules are often generated during the rule decoding stage, the LLM selects the optimal terminal symbol $\omega$ from $\Omega_\mathrm{candidate}$.}
The $\omega$ that maximizes the following probability is then chosen:
\begin{align}
    P_\mathrm{LLM}(\omega |(x^{(i)}, G[y^{(i)}], y^{(i)})_{i=1}^N, x, \hat{G}[y], \hat{y}_\mathrm{valid}, \Omega_\mathrm{candidate}).
\end{align}
The selected candidate is appended to the end of $\hat{y}_\mathrm{valid}$, and the LLM generates the remaining part of the game description $\hat{y}$.
Next, the $y'$ that maximizes the following probability is generated:
\begin{align}
    P_\mathrm{LLM}(y' |(x^{(i)}, G[y^{(i)}], y^{(i)})_{i=1}^N, x, \hat{G}[y], \hat{y}_\mathrm{valid} + \omega).
\end{align}
The generated $\hat{y}$ updates $\hat{y}$ from the previous step.
This process is repeated until the entire generated $\hat{y}$ becomes parsable by the Earley parser or until a predetermined number of updates is reached.

\lstset{
    language={},
    basicstyle=\ttfamily\tiny,
    keywordstyle=\color{black},
    stringstyle=\color{black},
    commentstyle=\color{black},
    numbers=left,
    numberstyle=\tiny\color{gray},
    frame=none,
    breaklines=true,
    showstringspaces=false,
    tabsize=2,
    escapeinside={(*@}{@*)}
}

\section{Experiments}
\subsection{Datasets}
\review{The game descriptions we use for evaluation are obtained from the publicly available Ludii website~\cite{ludii_portal}}.
In Ludii, each game is assigned a category (\emph{e.g.}, ``Tic-Tac-Toe'' is categorized under ``Board/Space/Line'', and ``Tower of Hanoi'' under ``Puzzle/Planning'').
From the same category, one test example and three demonstration examples are extracted as a single instance, \review{leveraging the known benefit of using semantically similar examples to improve ICL performance~\cite{icl_demo}.}
Note that only examples where the token length of the game description $y$ is 300 or less are used.
The token length is calculated using the tokenizer of Llama3-8B-Instruct~\cite{llama3}.
The evaluation dataset is constructed from 100 randomly selected instances from all categories, and the same set of instances is used in all evaluations.
Each example consists of a natural language instruction $x$, a Ludii game description $y$, and the minimal grammar $G[y]$ required to construct $y$.
The instruction $x$ is composed of the metadata ``Description'' and ``Rules'' provided by the Ludii game system (see Fig.~\ref{fig:gdg_example} for an example).
\review{To improve the generality of the dataset, we use game descriptions where the unique functions defined within each game are expanded as $y$.}
The extended version is instantiated with primitive options and rulesets to avoid using the Ludii game system's meta-language features (definitions, options, rulesets, ranges, constants, etc.).
\final{A similar process is adopted in \cite{gavel}.}
For more details of the Ludii language, please refer to the Ludii language reference~\cite{ludii_language_reference}.
The grammar $G[y]$ is automatically extracted by the parser using the Lark library~\cite{lark_parser}, which is explained in the next subsection.

\subsection{\review{Methods}}
We compare the following methods:
\begin{itemize}
    \item \review{\textbf{Random}: A baseline method that generates the grammar $\hat{G}[y]$ necessary for constructing the game description $y$. Based on $\hat{G}[y]$, the method randomly samples the next expressions following ``(game'' to generate the game description.}
    \item \review{\textbf{Game Description Generation (GDG)}: A baseline method that directly predicts the game description $y$ from the demonstration examples and the query $x$ without using the predicted grammar $\hat{G}[y]$.}
    \item \review{\textbf{Grammar-based Game Description Generation (GGDG, ours)}: Our proposed method, which first generates the grammar $\hat{G}[y]$ required to construct the game description $y$ and subsequently generates $\hat{y}$ based on $\hat{G}[y]$. This method employs both rule decoding and game description decoding.}
    \item \review{\textbf{SFT+GDG}: A baseline method that predicts the game description from a query using an LLM with supervised fine-tuning (SFT). For SFT training data, pairs of query $x$ and game description $y$ are created from games available on the Ludii portal~\cite{ludii_portal} that are not included in the evaluation dataset.}
    \item \review{\textbf{SFT+GGDG (ours)}: A method that combines our proposed GGDG with SFT. For rule decoding, an LLM with SFT applied to query $x$ and grammar $G[y]$ pairs is used. For game description decoding, the same LLM model as in SFT+GDG is used, trained with SFT on query $x$ and game description $y$ pairs. }
\end{itemize}
\review{Since the model acquires knowledge of Ludii through SFT, demonstration examples are not used, \ie, zero-shot inference.}
% \begin{itemize}
%     \item Game Description Generation (GDG): Predicts the game description $y$ directly from the demonstration examples and the query $x$ without using the predicted grammar $\hat{G}[y]$.
%     \item Grammar-based Game Description Generation (GGDG): Generates the grammar $\hat{G}[y]$ required to construct the game description $y$, and then generates $\hat{y}$ based on $\hat{G}[y]$.
%     \item GGDG~+~Rule Decoding (RD): Uses the proposed rule decoding in addition to GGDG.
%     \item GGDG~+~RD~+~Game Description Decoding (GDD): Uses both the proposed rule decoding and game description decoding in addition to GGDG.
% \end{itemize}
% In addition to these, for GGDG and GGDG~+~GDD, we also compare the methods that use the oracle grammar $G[y]$ obtained from the ground truth with a parser based on grammar $G$ instead of the generated $\hat{G}[y]$.

\subsection{Implementation Details}
We use a parser built with a Python library called Lark~\cite{lark_parser} to extract the grammar rules of the Ludii game description.
The Ludii grammar in the Lark parser employs the rules listed in the Ludii language reference.
In our grammar-based iterative decoding, we set an upper limit on the number of iterations to suppress the number of LLM calls.
\review{Based on the ablation study in Sec.~\ref{subsec:ablation}, the iteration limit is set to 20 for rule decoding and 10 for game description decoding.}
Our prompts for grammar-based iterative decoding are shown in Fig~\ref{fig:prompts}.

We use the Llama3-8B-Instruct~\cite{llama3} model as our LLM.
\review{We chose an open-source LLM to ensure research reproducibility. 
Unlike commercial APIs, our approach is not affected by API changes or model updates.}
Llama3 is the latest series of open-source LLMs provided by Meta.
The Llama3 series includes pre-trained models with 8B and 70B parameters.
Compared to other models of similar size, the Llama3 series demonstrates superior performance across various benchmarks.
To ensure that our framework is practical in environments with limited computational resources, such as local deployments, we select the smaller 8B model.
% Additionally, we use the instruction fine-tuning~\cite{instruct_gpt, dpo} version to make the model accurately respond to the various instructions used in our framework.
Since the Llama3 models are pre-trained to accept sequences of up to 8,192 tokens, it is necessary to keep the prompts within this context length.
To address this, we set the number of demonstration examples to three.
We use two NVIDIA RTX 6000 Ada GPUs for our experiments.

\review{For SFT models, we apply SFT to Llama3-8B-Instruct using low-rank adaptation (LoRA)~\cite{lora}.
We set the SFT sequence length to 4096, LoRA alpha and $r$ to 16, and train the model for 3 epochs with a learning rate of 1e-4 and a warmup ratio of 0.03.}

\begin{table*}[t]
    \centering
    \caption{\review{\textbf{Comparison with baseline methods.} The best results are in \textbf{bold}.}}
    \begin{tabular}{lrrrr}
    \toprule
    Method & Compilability$\uparrow$ & Functionality$\uparrow$ & ROUGE$\uparrow$ & \makecell{Normalized \\ Concept Distance$\downarrow$} \\
    \midrule
    Random & 1.3$\pm$0.7 & 1.0$\pm$0.6 & 9.8$\pm$0.8 & 0.97$\pm$0.02\\
    GDG & 27.0$\pm$1.2 & 26.3$\pm$0.7 & 63.5$\pm$0.6 & 0.75$\pm$0.01 \\
    GGDG~(ours) & 64.0$\pm$1.5 & 56.7$\pm$2.3 & 60.5$\pm$0.6 & 0.46$\pm$0.03 \\
    SFT+GDG & 59.7$\pm$2.4 & 58.0$\pm$0.6 & \textbf{64.0$\pm$0.5} & 0.44$\pm$0.01 \\
    SFT+GGDG~(ours) & \textbf{72.0$\pm$2.1} & \textbf{70.3$\pm$1.8} & 63.8$\pm$0.7 & \textbf{0.33$\pm$0.02} \\
    \bottomrule
  \end{tabular}
  \label{tab:main_results}
\end{table*}

\section{\review{Experimental Results}}
\label{sec:experimental_results}
\subsection{Evaluation Metrics}
To evaluate the generated game descriptions, we use the following metrics:
\begin{itemize}
    \item \review{\textbf{Compilability}: The proportion of games that can be parsed and compiled by the Ludii game engine. If a game cannot be compiled, it is not evaluated in the functionality metric. The score is normalized from 0 to 100.}
    \item \review{\textbf{Functionality}: \final{The proportion of playable games. Although a game may compile without errors, it is still considered non-functional if conditions make it unplayable—such as when piece movements rely on undefined positions. The score is normalized from 0 to 100.}}
    \item \textbf{ROUGE}~\cite{rouge}: \review{This is an evaluation metric commonly used in program synthesis to measure the degree of linguistic match between the generated game description and the ground truth game description~\cite{code_generation_survey}.} It ranges from 0 to 100, with higher values indicating a greater degree of match. The calculation of this metric does not consider syntactic correctness but instead focuses on the similarity between texts. We use the F1 score of ROUGE-L. This value is calculated for each piece of test data, and the average of all data is reported.
    \item \final{\textbf{Normalized Concept Distance (NCD)}: NCD measures how closely a predicted game matches its ground truth in Ludii. Following \cite{board_concept, measuring}, games are represented as concept-value vectors derived from semantic features and from behavioral data obtained through automated playouts using a random policy. For example, these values include attributes such as the proportion of turns with at least one legal move and the proportion of the board used at least once. The cosine distance between these vectors gives NCD. Similar to \cite{gavel}, we run 50 playouts for the ground truth and 10 for the predicted games due to computational costs. Since non-functional games cannot compute their concept distance, their distance is set to 1.0. NCD is averaged over all test data, serving as a quality measure for game description generation.}
    % \item Grammar Parsability: The proportion of generated game descriptions that can be parsed by the Lark parser, which is based on the Ludii grammar $G$.% (not $\hat{G}[y]$).
    % \item System Parsability: The proportion of generated game descriptions that can be read by the Ludii game system. Descriptions that are grammatically correct as per the Lark parser but lack any of the necessary functions for a game can not be read by the Ludii game system.
    % \item Playability: The proportion of generated game descriptions that require a reasonable number of player actions per trial. A trial is considered successful if it fulfills the specified termination conditions in at least 5 out of 10 attempts. In particular, we consider the game to be playable if, by continuously choosing legal actions, the number of actions necessary to reach a termination condition should range between 2 and 1,800. This metric evaluates whether the generated game has a reasonable length of play experience. For a game description to satisfy the Playability criteria, it also has to satisfy the Grammar Parsability and System Parsability criteria. We use the implementation~\cite{ludii_portal} in the Ludii game system for this metric.
    % \item Levenshtein Distance: The edit distance between the generated game description and the ground truth.
    % \item Exact Match: The proportion of generated game descriptions that exactly match the ground truth.
\end{itemize}
\review{We conducted experiments with 3 different seeds, and each value in the results shows the mean and standard error across the seeds.}

\subsection{Comparison with Baseline Methods}
\review{Table~\ref{tab:main_results} shows the comparison results with baseline methods.
GGDG outperforms GDG with Compilability~+37.0, Functionality~+30.4, and NCD~-0.29, demonstrating that our rule decoding and game description decoding effectively improve the grammatical accuracy of generated game descriptions.
In ROUGE, GGDG scores -3.0 lower than GDG.
Since ROUGE evaluates the level of reproduction in linguistic expressions, it may undervalue different description methods generated by GGDG that provide the same playable experience.
For example, the ending rule for Tic-tac-toe can be defined as either ``win by aligning three pieces'' or ``lose when the opponent aligns three pieces'' – while these expressions have the same meaning as rules, they would result in a lower ROUGE score.}

\review{While GGDG shows a +4.3 advantage in Compilability compared to SFT+GDG, it performs worse in Functionality (-1.3), ROUGE (-3.5), and NCD (-0.02).
This suggests that although GGDG can generate grammatically accurate game descriptions, it still faces challenges compared to SFT in satisfying functionality requirements and reproducing the game quality of the ground truth.
SFT+GGDG achieves the best scores across Compilability (72.0), Functionality (70.3), and NCD (0.33) metrics, demonstrating that SFT has a complementary effect on GGDG.}

\review{Regarding the baseline results, random generation methods can hardly produce grammatically valid game descriptions.
While GDG shows high ROUGE scores, it performs poorly on other metrics.
This suggests that although LLMs alone can generate game descriptions that are superficially similar in linguistic expression, without accessing Ludii's grammar through our decoding methods, they struggle to generate grammatically and functionally accurate game descriptions.
Additionally, SFT+GDG shows improved performance across all metrics compared to GDG, indicating that providing the model with Ludii knowledge through SFT further enhances grammatical accuracy.}

\begin{table*}[t]
    \centering
    \caption{\review{\textbf{Ablation study of our proposed method.} ``Oracle Grammar'' indicates cases where grammar $G[y]$ extracted from game description $y$ is used instead of grammar $\hat{G}[y]$ predicted by the LLM. ``Grammar'' indicates the inclusion of grammar in the query during game description generation. The best results are in \textbf{bold}.}}
    \begin{tabular}{lcccrrrr}
    \toprule
    Method & Grammar & \makecell{Rule \\ Decoding} & \makecell{Game Description \\ Decoding} & Compilability$\uparrow$ & Functionality$\uparrow$ & ROUGE$\uparrow$ & \makecell{Normalized \\ Concept Distance$\downarrow$} \\
    \midrule
    GDG & & & & 27.0$\pm$1.2 & 26.3$\pm$0.7 & 63.5$\pm$0.6 & 0.75$\pm$0.01 \\
    GGDG w/o RD, GDD & $\checkmark$ & & & 48.3$\pm$0.3 & 43.3$\pm$1.2 & 61.6$\pm$0.1 & 0.59$\pm$0.01 \\
    GGDG w/o GDD & $\checkmark$ & $\checkmark$ & & 52.0$\pm$3.6 & 47.3$\pm$2.3 & 61.3$\pm$0.2 & 0.55$\pm$0.02 \\
    GGDG & $\checkmark$ & $\checkmark$ & $\checkmark$ & \textbf{64.0$\pm$1.5} & 56.7$\pm$2.3 & 60.5$\pm$0.6 & \textbf{0.46$\pm$0.03} \\
    \midrule
    \multicolumn{3}{l}{\footnotesize{\textit{Oracle Grammar}}} \\
    \addlinespace[3pt]
    GGDG w/o RD, GDD & $\checkmark$ & & & 54.0$\pm$1.7 & 49.3$\pm$0.7 & 61.7$\pm$0.2 & 0.53$\pm$0.01 \\
    GGDG w/o RD & $\checkmark$ & & $\checkmark$ & 62.0$\pm$1.5 & \textbf{57.3$\pm$2.2} & \textbf{64.5$\pm$0.5} & \textbf{0.46$\pm$0.02} \\
    \bottomrule
  \end{tabular}
  \label{tab:component_results}
\end{table*}

\subsection{Ablation Study}
\label{subsec:ablation}

\noindent \review{\textbf{GGDG Components.} We conduct an ablation study on GGDG and summarize the results in Tab.~\ref{tab:component_results}.
We confirm that adding more techniques improves the scores for Compilability, Functionality, and NCD.
Among the improvements from GDG to GGDG, grammar accounts for the largest share, comprising 57.6\% of Compilability, 55.9\% of Functionality, and 55.2\% of NCD improvements.
This is followed by game description decoding, which accounts for 32.4\% of Compilability, 30.9\% of Functionality, and 31.0\% of NCD improvements, demonstrating their substantial impact.
Furthermore, the improvement in Functionality and ROUGE scores when using oracle grammar suggests that there is still room for improvement in grammar generation through rule decoding.}

\begin{table}[t]
    \setlength{\tabcolsep}{1mm}
    \centering
    \caption{\review{\textbf{Comparison of the iteration limits for Rule Decoding.} Game Description Decoding is not used (i.e., GGDG w/o GDD). The best results are in \textbf{bold}.}}
    \begin{tabular}{crrrrr}
    \toprule
    Iteration & Compilability$\uparrow$ & Functionality$\uparrow$ & ROUGE$\uparrow$ & NCD$\downarrow$ \\
    \midrule
    10 & 50.0$\pm$1.0 & 44.7$\pm$2.0 & 61.5$\pm$0.1 & 0.58$\pm$0.02 \\
    20 & \textbf{51.3$\pm$3.5} & \textbf{46.7$\pm$2.4} & 61.3$\pm$0.2 & \textbf{0.55$\pm$0.02} \\
    30 & 47.7$\pm$2.7 & 44.0$\pm$3.5 & \textbf{62.1$\pm$0.2} & 0.58$\pm$0.03 \\
    \bottomrule
  \end{tabular}
  \label{tab:rd_iteration_results}
\end{table}

\begin{table}[t]
    \setlength{\tabcolsep}{1mm}
    \centering
    \caption{\review{\textbf{Comparison of the iteration limits for Game Description Decoding.} The number of iterations for Rule Decoding is fixed at 20. The best results are in \textbf{bold}.}}
    \begin{tabular}{crrrrr}
    \toprule
    Iteration & Compilability$\uparrow$ & Functionality$\uparrow$ & ROUGE$\uparrow$ & NCD$\downarrow$ \\
    \midrule
    5 & 60.0$\pm$3.2 & 53.7$\pm$2.8 & \textbf{61.1$\pm$0.2} & 0.50$\pm$0.02 \\
    10 & 64.0$\pm$1.5 & 56.7$\pm$2.3 & 60.5$\pm$0.6 & \textbf{0.46$\pm$0.03} \\
    20 & \textbf{64.7$\pm$0.3} & \textbf{57.3$\pm$1.9} & 60.6$\pm$0.2 & 0.47$\pm$0.01 \\
    \bottomrule
  \end{tabular}
  \label{tab:gdd_iteration_results}
\end{table}

\noindent \review{\textbf{Iteration Limit for Decoding.} We investigate the impact of iteration limits in our decoding approach.
First, Tab.~\ref{tab:rd_iteration_results} shows the results of rule decoding.
When the iteration limit is set to 30, we find that the Compilability and Functionality scores are the lowest.
This may be because increasing the iteration limit leads to the inclusion of unnecessary rules, potentially degrading performance.
When the iteration limit is set to 20, Compilability, Functionality, and NCD scores show the best results.
Therefore, we set the iteration limit to 20 for subsequent experiments.}

\review{Tab.~\ref{tab:gdd_iteration_results} shows the results of game description decoding. When the iteration limit is set to 5, we find that the Compilability and Functionality scores are the lowest.
This is likely due to insufficient iterations for improving the game description.
When the iteration limit is set to 10 or 20, we observe minimal differences across most metrics.
When the iteration limit is 10, NCD is lowest at 0.46 and the computational cost is also low, therefore we set the iteration limit to 10 in other experiments.}

\begin{table}[t]
    \setlength{\tabcolsep}{1mm}
    \centering
    \caption{\review{\textbf{Comparison of game categories in demonstration examples.} ``Same'' category indicates that examples are from the same category as the test instance, while ``Cross'' category indicates that they are from different categories. The best results are in \textbf{bold}.}}
    \begin{tabular}{lrrrrr}
    \toprule
    Method & Compilability$\uparrow$ & Functionality$\uparrow$ & ROUGE$\uparrow$ & NCD$\downarrow$ \\
    \midrule
    \multicolumn{3}{l}{\footnotesize{\textit{Cross Category}}} \\
    \addlinespace[3pt]
    GDG & 5.3$\pm$0.3 & 3.7$\pm$0.7 & 46.0$\pm$0.0 & 0.97$\pm$0.00 \\
    GGDG & 36.0$\pm$0.6 & 24.6$\pm$0.3 & 42.3$\pm$0.1 & 0.79$\pm$0.00 \\
    \midrule
    \multicolumn{3}{l}{\footnotesize{\textit{Same Category}}} \\
    \addlinespace[3pt]
    GDG & 27.0$\pm$1.2 & 26.3$\pm$0.7 & \textbf{63.5$\pm$0.6} & 0.75$\pm$0.01 \\
    GGDG & \textbf{64.0$\pm$1.5} & \textbf{56.7$\pm$2.3} & 60.5$\pm$0.6 & \textbf{0.46$\pm$0.03} \\
    \bottomrule
  \end{tabular}
  \label{tab:demo_category_results}
\end{table}

\noindent \review{\textbf{Game Category of Demonstration Examples.} We investigate how the category of demonstration examples affects game performance.
In our proposed method, we use instances from the same category as the test instances for demonstration examples (Same Category).
We compare this with using instances from different categories (Cross Category), and summarize the results in Tab.~\ref{tab:demo_category_results}.
In Cross Category, demonstration examples are randomly selected from all games except for the category of the test instance.
We find that obtaining demonstration examples from the same category as test instances significantly improves performance across all evaluation metrics. }

\subsection{Impact of Game Characteristics and Model Configurations}
\begin{table}[t]
    \setlength{\tabcolsep}{1mm}
    \centering
    \caption{\review{\textbf{Comparison of test game lengths.} \textit{300 - 500} indicates that the evaluation is performed on games with game description token lengths between 300 and 500.}}
    \begin{tabular}{lrrrrr}
    \toprule
    Method & Compilability$\uparrow$ & Functionality$\uparrow$ & ROUGE$\uparrow$ & NCD$\downarrow$ \\
    \midrule
    \multicolumn{3}{l}{\footnotesize{\textit{0 - 300}}} \\
    \addlinespace[3pt]
    GDG & 27.0$\pm$1.2 & 26.3$\pm$0.7 & 63.5$\pm$0.6 & 0.75$\pm$0.01 \\
    GGDG & 64.0$\pm$1.5 & 56.7$\pm$2.3 & 60.5$\pm$0.6 & 0.46$\pm$0.03 \\
    \midrule
    \multicolumn{3}{l}{\footnotesize{\textit{300 - 500}}} \\
    \addlinespace[3pt]
    GDG & 41.0$\pm$2.1 & 39.0$\pm$1.0 & 59.7$\pm$0.2 & 0.68$\pm$0.01 \\
    GGDG & 54.3$\pm$2.2 & 48.7$\pm$1.3 & 59.4$\pm$0.5 & 0.59$\pm$0.01 \\
    \midrule
    \multicolumn{3}{l}{\footnotesize{\textit{500 - 1000}}} \\
    \addlinespace[3pt]
    GDG & 36.0$\pm$2.3 & 32.7$\pm$2.6 & 49.9$\pm$0.3 & 0.74$\pm$0.02 \\
    GGDG & 33.3$\pm$1.2 & 29.7$\pm$0.9 & 47.3$\pm$0.6 & 0.77$\pm$0.00 \\
    \bottomrule
  \end{tabular}
  \label{tab:game_length_results}
\end{table}

\noindent \review{\textbf{Game Description Length.} We investigated the impact of game description length on performance.
We compared three groups based on token length: 0-300, 300-500, and 500-1,000.
Token lengths were calculated using the Llama-3-8B-Instruct tokenizer.
The results are summarized in Tab.~\ref{tab:game_length_results}.
For 0-300 and 300-500 tokens, \final{GGDG outperforms GDG} in Compilability, Functionality, and NCD metrics, demonstrating its ability to generate more grammatically accurate game descriptions.
Meanwhile, GGDG showed declining performance across all metrics as token length increased, dropping from 64.0 to 33.3, Functionality from 56.7 to 29.7, ROUGE from 60.5 to 47.3, and NCD worsening from 0.46 to 0.77.
For 500-1,000 tokens, GDG slightly outperformed GGDG across all metrics.
These results suggest that GGDG struggles with long game descriptions.
We discuss this limitation in Sec~\ref{sec:limitation}.}

\begin{table}[t]
    \setlength{\tabcolsep}{1mm}
    \centering
    \caption{\review{\textbf{Comparison of test instance categories.}}}
    \begin{tabular}{lrrrrr}
    \toprule
    Method & Compilability$\uparrow$ & Functionality$\uparrow$ & ROUGE$\uparrow$ & NCD$\downarrow$ \\
    \midrule
    \multicolumn{3}{l}{\footnotesize{\textit{board/race}}} \\
    \addlinespace[3pt]
    GDG & 10.0$\pm$5.8 & 10.0$\pm$5.8 & 58.2$\pm$0.2 & 0.90$\pm$0.06 \\
    GGDG & 60.0$\pm$5.8 & 60.0$\pm$5.8 & 58.8$\pm$0.2 & 0.42$\pm$0.05 \\
    \midrule
    \multicolumn{3}{l}{\footnotesize{\textit{board/sow}}} \\
    \addlinespace[3pt]
    GDG & 38.9$\pm$5.6 & 38.9$\pm$5.6 & 78.8$\pm$1.0 & 0.75$\pm$0.06 \\
    GGDG & 55.6$\pm$5.6 & 55.6$\pm$5.6 & 81.3$\pm$0.9 & 0.51$\pm$0.06 \\
    \midrule
    \multicolumn{3}{l}{\footnotesize{\textit{puzzle}}} \\
    \addlinespace[3pt]
    GDG & 7.4$\pm$1.9 & 7.4$\pm$1.9 & 51.3$\pm$0.5 & 0.98$\pm$0.02 \\
    GGDG & 61.1$\pm$6.4 & 59.3$\pm$6.7 & 52.1$\pm$0.7 & 0.57$\pm$0.12 \\
    \midrule
    \multicolumn{3}{l}{\footnotesize{\textit{board/space/line}}} \\
    \addlinespace[3pt]
    GDG & 26.9$\pm$0.9 & 24.4$\pm$0.5 & 64.2$\pm$0.3 & 0.78$\pm$0.00 \\
    GGDG & 68.7$\pm$0.9 & 63.7$\pm$1.3 & 62.4$\pm$0.4 & 0.42$\pm$0.02 \\
    \midrule
    \multicolumn{3}{l}{\footnotesize{\textit{board/war}}} \\
    \addlinespace[3pt]
    GDG & 57.7$\pm$4.4 & 57.7$\pm$4.4 & 71.0$\pm$0.1 & 0.45$\pm$0.04 \\
    GGDG & 71.8$\pm$7.1 & 62.8$\pm$9.0 & 71.8$\pm$0.3 & 0.41$\pm$0.09 \\
    \bottomrule
  \end{tabular}
  \label{tab:test_category_results}
\end{table}

\noindent \review{\textbf{Game Category of Test Games.} We investigate the impact of game categories of test instances.
We compare GDG and GGDG across five categories: racing games (board/race), mancala games (board/sow), puzzle games (puzzle), line games (board/space/line), and war games, including capture games (board/war).
The demonstration examples are from the same category as the test instances.
We use instances with game description token lengths of 300 or less as test games. }

\review{The results are summarized in Tab.~\ref{tab:test_category_results}.
Across all categories, our proposed GGDG outperforms GDG in terms of Compilability, Functionality, and NCD metrics, which is consistent with our previous experimental results.
For ROUGE scores, while the superior method varies by category, the difference between GDG and GGDG remains within 2 points across most categories.}

\review{In comparisons across categories, GDG's performance shows significant variation, with the puzzle category showing the lowest scores (Compilability/Functionality: 7.4, NCD: 0.98) for GDG and the largest improvement margin (+53.7/+51.9 and -0.41 respectively) when using GGDG.
Conversely, the board/war categories showed GDG's highest performance levels (Compilability/Functionality: 57.7, NCD: 0.45) and the smallest margin of improvement with GGDG (+14.1/+5.1 and -0.04 respectively).
This variation may be attributed to games in the puzzle category having more complex logic compared to other categories, making it difficult for LLMs to learn grammatically correct patterns from demonstration examples, or possibly due to limited knowledge about puzzle category games obtained during pre-training.
Meanwhile, GGDG showed smaller performance variations between categories compared to GDG, with scores ranging from 55.6 to 71.8 for Compilability, 55.6 to 63.7 for Functionality, and 0.41 to 0.57 for NCD, demonstrating more stable performance.}

\begin{table}[t]
    \setlength{\tabcolsep}{1mm}
    \centering
    \caption{\review{\textbf{Comparison of LLM models.} The best results are in \textbf{bold}.}}
    \begin{tabular}{lrrrrr}
    \toprule
    Method & Compilability$\uparrow$ & Functionality$\uparrow$ & ROUGE$\uparrow$ & NCD$\downarrow$ \\
    \midrule
    \multicolumn{3}{l}{\footnotesize{\textit{Llama-3.2-3B-Instruct}}} \\
    \addlinespace[3pt]
    GDG & 18.7$\pm$0.7 & 17.3$\pm$0.3 & 59.2$\pm$0.3 & 0.83$\pm$0.00 \\
    GGDG & 28.0$\pm$1.5 & 25.7$\pm$2.0 & 55.4$\pm$0.1 & 0.76$\pm$0.02 \\
    \midrule
    \multicolumn{3}{l}{\footnotesize{\textit{Llama-3-8B-Instruct}}} \\
    \addlinespace[3pt]
    GDG & 27.0$\pm$1.2 & 26.3$\pm$0.7 & 63.5$\pm$0.6 & 0.75$\pm$0.01 \\
    GGDG & 64.0$\pm$1.5 & 56.7$\pm$2.3 & 60.5$\pm$0.6 & 0.46$\pm$0.03 \\
    \midrule
    \multicolumn{3}{l}{\footnotesize{\textit{gpt-4o}}} \\
    \addlinespace[3pt]
    GDG & 27.7$\pm$0.3 & 27.7$\pm$0.3 & 67.7$\pm$0.1 & 0.74$\pm$0.00 \\
    GGDG & \textbf{70.7$\pm$0.9} & \textbf{59.3$\pm$0.9} & \textbf{68.2$\pm$0.0} & \textbf{0.44$\pm$0.01} \\
    \bottomrule
  \end{tabular}
  \label{tab:llm_results}
\end{table}

\noindent \review{\textbf{LLM Models.} We compare the performance of different LLM models. We primarily used Llama-3-8B-Instruct in our experiments and compare its performance to that of Llama-3.2-3B-Instruct, a smaller LLM model, and gpt-4o, a more powerful LLM model. The results are summarized in Tab.~\ref{tab:llm_results}. GGDG improves Compilability, Functionality, and NCD scores over GDG across all models. With gpt-4o, GGDG also improves the ROUGE score. Comparing across models, the 8B model shows significant improvements in all evaluation metrics compared to the 3B model. For example, with GGDG, upgrading from the 3B model to the 8B model improves Compilability by +36.0, Functionality by +31.0, ROUGE by +5.1, and NCD by -0.30. gpt-4o achieves the highest scores across all evaluation metrics. The improvements over the 8B model are +6.7 in Compilability, +2.6 in Functionality, +7.7 in ROUGE, and -0.02 in NCD. We observe that the improvements in Compilability, Functionality, and NCD are larger when moving from the 3B model to the 8B model compared to moving from the 8B model to gpt-4o. This suggests that while increasing model size brings significant performance improvements, the room for improvement becomes limited between already high-performing models. Therefore, if higher performance is desired, selecting powerful models like gpt-4o is effective. In contrast, if there are resource constraints, such as API costs or computational resource limitations, the 8B model can be considered a balanced choice between performance and cost.}

\subsection{Qualitative Results}
\begin{figure*}[t]
    \centering
    \includegraphics[width=\linewidth]{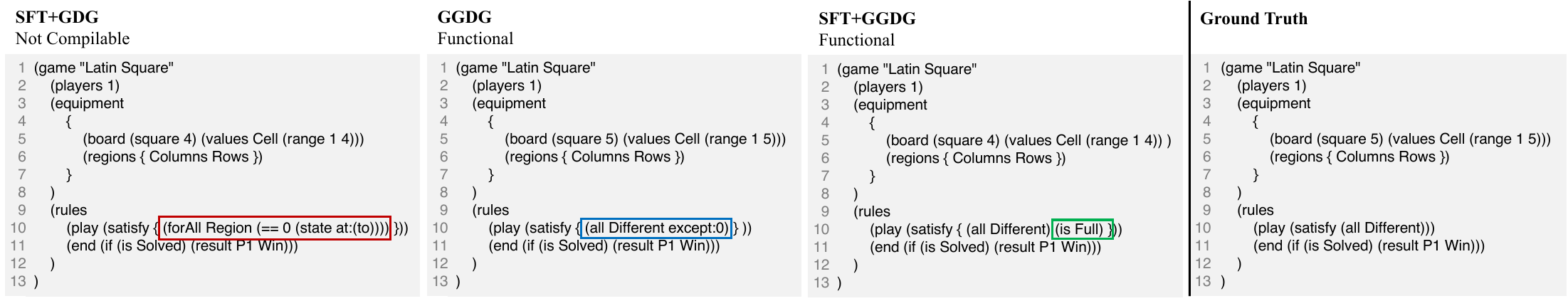}
    \caption{\review{\textbf{Comparison of generation results with baseline methods for Latin Square.} Since the board size is not specified in the query, it varies depending on the generated results. The ground truth is set to size 5 as an example.}}
    \label{fig:latin_square}
\end{figure*}

\noindent \review{\textbf{Comparison with Baseline Methods.}
We conduct a qualitative analysis comparing SFT+GDG, which is the most promising among the baseline methods, with our proposed methods, GGDG and SFT+GGDG.
Figure~\ref{fig:latin_square} shows the generation results for Latin Square.
Latin Square is a puzzle where players place numbers in an n$\times$n grid such that the same number does not repeat in each row and column.}

\review{The result of SFT+GDG is neither compilable nor functional due to the part shown in the red frame: ``(forAll Region (== 0 (state at:(to)))''.
This issue stems mainly from two grammatical problems.
First, while the Ludii grammar defines that a ``forAll'' clause must be followed by a ``puzzleElementType'' (either ``Cell'', ``Edge'', ``Vertex'', or ``Hint''), an undefined terminal symbol, ``Region'', is used.
Second, the expression ``(== 0 (state at:(to)))'' cannot be parsed because the operator ``=='' is not defined in the Ludii grammar.}

\review{In contrast, GGDG's result is both compilable and functional, generating a game description almost identical to the ground truth.
In particular, it shows significant improvement in that the result contains no grammatical errors.
However, in the blue frame, ``except:0'' (a constraint requiring all values to be different except for index 0) is added after ``all Different'', and this rule is not included in the game rules of Latin Square.}

\review{Furthermore, the SFT+GGDG result shows more improvement and is consistent with the Latin Square game rules.
The ``is Full'' rule, which requires all cells to be filled, is redundant as it is already included in the content of the ``all Different'' rule, but it does not contradict the Latin Square rules.
These results suggest that combining SFT and GGDG improves performance.}

\begin{figure*}[t]
    \centering
    \includegraphics[width=\linewidth]{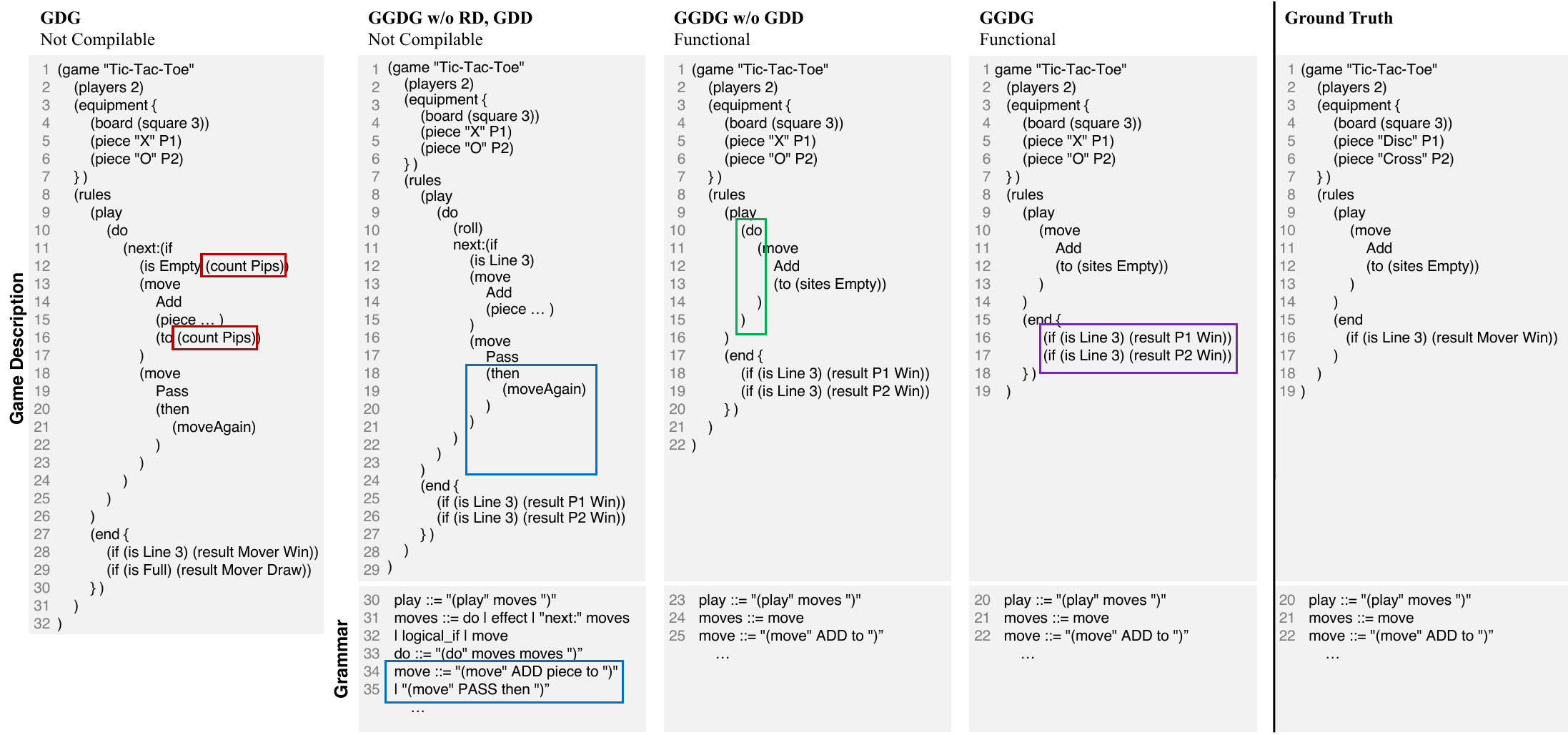}
    \caption{\review{\textbf{Impact of GGDG components on generation results for Tic-Tac-Toe.} Components are added from left to right, with the ground truth on the far right. The middle row shows the game description, and the bottom row shows the results of grammar generation for only the methods that performed grammar generation. Since the grammar results are highly redundant, only the important parts are shown.}}
    \label{fig:components_qualitative_results}
\end{figure*}

\noindent \review{\textbf{GGDG Components.} We investigate the effects of GGDG components through qualitative analysis.
Figure~\ref{fig:components_qualitative_results} shows the generation results for Tic-Tac-Toe.
The GDG results are not compilable due to the ``count Pips'' shown in the red frame.
The appearance of the term ``Pips'' occurs because the LLM is influenced by the dice in the Tic-Tac-Die game from the demonstration examples.
Tic-Tac-Die is explained as follows: ``Tic-Tac-Die is played similarly to Tic-Tac-Toe except that players roll a D9 dice each turn to dictate where they move (dice pips show the cell index to move to).''}

\review{While the pips disappear in the results of GGDG~w/o~RD,~GDD, they are still not compilable.
This is due to the extra clause about ``PASS'' in the blue frame.
This occurs because the predicted ``move'' grammar contains a ``PASS'' clause not included in the ground truth.}

\review{The grammar results of GGDG~w/o~GDD show that by adding rule decoding, they match the ground truth grammar.
The generated game description is functional.
However, this description includes ``do,'' which is not in the predicted grammar.
The presence of this term, which appears in the Tic-Tac-Die demonstration example, suggests that the LLM is mimicking it.}

\review{In the GGDG results, adding game description decoding reduces the occurrence of expressions not defined in the grammar, such as ``do''.
This game description is functional and closest to the ground truth among the compared methods.
However, since the condition ``(is line 3)'' is applied equally to player1 and player2 as shown in the purple frame, it cannot accurately determine which player has won.
This suggests there is still room for improvement in GGDG.}

\begin{figure*}[t]
    \centering
    \includegraphics[width=\linewidth]{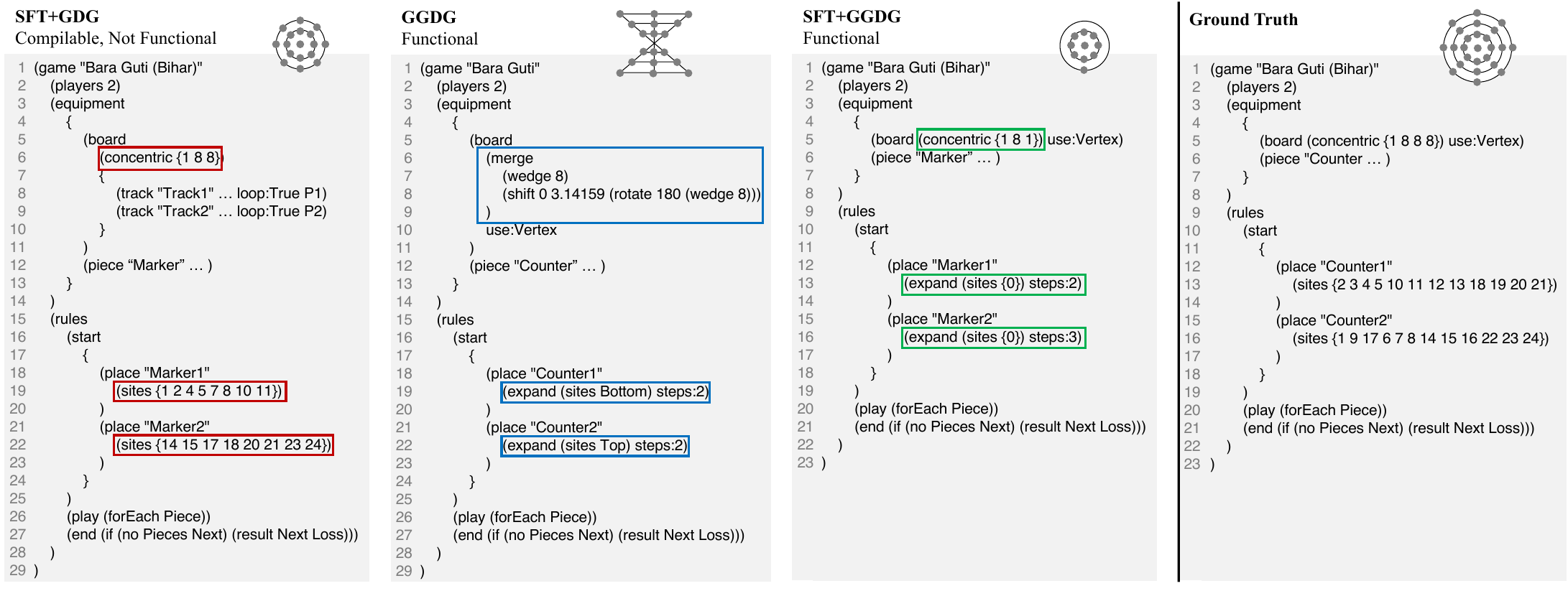}
    \caption{\review{\textbf{Comparison of generation results for Bara Guti (Bihar) as a failure example.} Some parts of the game description are omitted due to space constraints. The board appears in the upper right of each game description. Due to space constraints, GGDG's results display a wedge of 4 instead of 8.}}
    \label{fig:bara_quti}
\end{figure*}

\noindent \review{\textbf{Analysis of Failure Cases.} We examine the analysis results of Bara Guti (Bihar) as an example of a failure case, as shown in Fig.~\ref{fig:bara_quti}.
This game uses a board that has three concentric circles, with four diameters dividing it into eight equal sections, with counters placed in each section.
In analyzing the generated results, we note that "Markers" and "Counters" refer to the same type of piece.}

\review{While SFT+GDG generated compilable output, it did not produce functional results due to the parts highlighted in red in Fig.~\ref{fig:bara_quti}.
``concentric {1 8 8}'' specifies the number of cells in each concentric ring sequentially, defining a total of 17 cells.
However, this description becomes non-functional because lines 19 and 21 specify marker initial positions with indices of 17 or greater.
Additionally, the specified initial positions differ from the ground truth.}

\review{In contrast, while GGDG's generated results demonstrate functionality, the definitions of the board and counter initial positions differ from the ground truth.
Specifically, it generates a board by connecting two triangles on lines 19 and 22, and places counters at positions expanded two steps from the bottom using the expression ``expand (sites Bottom) steps:2'' (highlighted in blue).
This difference is presumed to be influenced by the existing demonstration example of Lau Kata Kati: ``(merge (wedge 4) (shift 0 3 (rotate 180 (wedge 4))))''.}

\review{Finally, while SFT+GGDG's results maintain functionality and return to the concentric circle board structure, the number of concentric circles, cell count, and Marker initial positions are incorrectly predicted, highlighted in green.
These limitations can be attributed to LLM’s lack of ability to understand spatial structures like boards~\cite{spatial_understanding}.
We discuss these limitations of spatial understanding in Sec~\ref{sec:limitation}.}

\section{Limitations and Discussion}
\label{sec:limitation}
\review{As shown in Tab.~\ref{tab:game_length_results}, GGDG's performance decreases with longer game descriptions.
It is known that LLM performance degrades with longer input sequences~\cite{long_context}.
We believe that GGDG struggles with processing longer game descriptions because its input token length is increased by the grammar, compared to GDG.
The development of LLMs capable of handling longer contexts is ongoing~\cite{longlm, efficientlong}, and newer LLM models may help mitigate this issue.}

\review{As shown in Fig.~\ref{fig:bara_quti}, it is difficult to compensate for LLMs' lack of spatial understanding capabilities through SFT.
It has been found that LLMs with 70B parameters demonstrate superior capabilities compared to models with 7B or 13B parameters in this regard~\cite{spatial_understanding}.
Using a larger LLM, such as the 70B parameter models, is a possible option to increase such capabilities.
%Therefore, applying SFT to large-parameter LLMs, such as 70B models, is one potential option.
However, it should be noted that SFT for large-parameter LLMs requires substantial computational resources, which can be problematic in other ways.}

Our proposed framework may have issues with inference time and cost because it repeats LLM inference multiple times.
\review{Specifically, the inference time per game averages 7.7 seconds for GDG and 143.9 seconds for GGDG.}
This issue can be mitigated by caching LLM responses, setting appropriate limits on the number of LLM inferences, and improving the prompts.
\review{Additionally, given recent advances in inference acceleration techniques~\cite{llm_accelarate}, these constraints are expected to be lessened as newer techniques are developed.}

The evaluation in this paper relies solely on automatically calculable metrics and does not include \review{human evaluations}.
\review{We believe that human evaluation could capture errors that automated evaluation cannot fully cover, potentially leading to further improvements and leads for future research directions.}

\section{Broader Impact}
The use of LLMs in the field of video games is associated with ethical issues related to sustainability, copyright, explainability, and biases~\cite{gallotta2024large}. 
In this section, we discuss the problems and their mitigation strategies within our proposed framework from these perspectives.

Our research raises concerns about the carbon footprint of LLMs. 
Our framework does not include training LLMs, and only LLMs’ inference impacts the environment. 
This issue can be mitigated by reducing the number of inference calls through caching LLM responses and providing more effective demonstrations. 
Additionally, using models that offer enhanced performance for the same computational load can further reduce environmental impact.

Our research is related to copyright issues concerning input and output data. 
It is common practice for LLMs to be trained using copyrighted data~\cite{gallotta2024large}. 
Since our framework does not include the training process, it does not directly cause this issue. 
However, users should consider the training data of LLMs when employing our framework. 
Our approach automatically generates game descriptions without human intervention, and these outputs may not be copyrighted.
Developing our framework into an interactive approach with human designers may allow the final output to qualify for copyright.

\review{The generation process of LLMs is opaque, and there are studies examining whether LLMs' black-box nature hinders the transparency of PCG~\cite{prompt-wrangling}. }
In our research as well, the process of each LLM call remains similarly unclear. 
However, our grammar-based iterative decoding method breaks down the generation process into multiple steps, which should enhance our understanding of it.

LLMs are trained on data collected from the internet, which introduces biases. 
Since game design tends to reflect the culture of its time and region, these biases negatively impact the design process when using LLMs. 
Our framework could potentially mitigate the issue by incorporating examples from various eras and regions in the demonstrations for LLMs. 
Ludii collects historically influential games from diverse regions and provides metadata on the regions where each game is rooted, which can contribute to mitigating this problem. 
\section{Conclusions}
\review{We have proposed a novel framework that combines large language models with game description languages for generating game descriptions from text.
Our approach integrates GDL grammar into the generation process, enabling the creation of structurally coherent game descriptions.
By introducing iterative refinement decoding methods specialized for both grammar generation and game description generation, we have seen improvements in the grammatical accuracy of game descriptions.
Extensive experimental results demonstrate that our framework is effective in improving grammatical accuracy within game description generation.
Future research may explore fine-tuning specialized LLMs for each subtask of rule and game description completion performed at each step of iterative refinement decoding, as well as the utilization of more efficient and powerful LLMs.
Additionally, it would be valuable to qualitatively validate the findings of our approach beyond the Ludii dataset through conducting user studies with actual humans.}

\section*{Acknowledgments}
This work was supported by JST, ACT-X Grant Number JPMJAX23CE, Japan.

\bibliographystyle{IEEEtran}
\bibliography{abrv,reference}

\vfill

\end{document}